# Knowledge-Driven Dimension Estimation from a Single Image

3D Asset Generation Technology for Digital Twin Construction

Hidenori Sakaniwa  Akihito Akai  Akihiko Hyodo

Data & Knowledge Management Research Department, Digital Infrastructure Innovation Center, Research & Development Group, Hitachi, Ltd., 1-280, Higashi-koigakubo, Kokubunji-shi, Tokyo, 185-8601, Japan

Abstract: In the verification of in-vehicle cameras, simulation technology using virtual spaces has advanced, enabling pre-evaluation of false detections and missed detections in various scenarios. However, discrepancies in the scale of the object being verified between the virtual and real environments can lead to a decrease in camera recognition performance. For traffic signs installed at high altitudes, distance measurement using LiDAR or stereo cameras is difficult, requiring size estimation from monocular images. This paper proposes a method for estimating the scale of an object by decomposing it into multiple structural elements and integrating external knowledge regarding design rules, geometric relationships, and conventional dimensions. Specifically, this method detects each component from a monocular image and estimates the size of each component by considering its structural relationships and dimensional consistency with surrounding elements. Furthermore, it generates a 3D asset of the object by reconstructing the estimated components. This method makes it possible to place 3D assets with a scale approximating the real environment within a digital twin space and is expected to contribute to improving the verification accuracy of in-vehicle cameras for autonomous driving in virtual environments.

## 1. Introduction

In the development of autonomous driving, large-scale verification is essential from the perspective of ensuring safety, and the importance of pre-evaluation using virtual environments such as digital twins is increasing. While virtual environments enable comprehensive simulation of verification scenarios that are difficult to reproduce in the real world, objects in virtual space may look similar but do not necessarily match the dimensions of the real world. To achieve accurate simulation, the dimensions of virtual objects must be strictly matched with those of the real world. For example, if the dimensions of road signs or information boards do not match those of the real world, there will be a difference between the distance at which a virtual camera can detect a sign and the distance at which a real-world camera can detect it, reducing the accuracy of the verification.

In recent years, data from sensors such as LiDAR and stereo cameras have enabled the construction of accurate point clouds and the creation of virtual environments whose metric dimensions closely match those of the physical world. However, LiDAR and stereo cameras mounted on vehicles are primarily designed to detect surrounding vehicles and pedestrians; consequently, even when leveraging existing datasets, they generally fail to capture elevated assets—such as overhead guide signs—with sufficient fidelity, making accurate dimensional reconstruction difficult. Moreover, individually measuring the sizes of the numerous elevated installations is impractical, and acquiring new high-elevation point clouds with additional sensors is costly. Thus, in the construction of digital twins, accurately and efficiently reproducing the dimensions of digital assets for elevated roadside infrastructure remains a persistent challenge. This paper focuses on the problem of estimating the dimensions of a target object from a single image when absolute dimensional information is unavailable or point cloud data is incomplete. To address this challenge, we propose a knowledge-driven framework that detects and recognizes the target object from an image and then estimates its dimensions by utilizing existing knowledge and design information with similar structures, based on the characteristics of its components and the dimensional correlations between those components.

## 2. Related Work

Digital twins have been widely studied as an integrated platform combining 3D models, sensing data, and simulations for urban development and mobility applications [1-3]. Previous research has mainly focused on linking real-world information to virtual spaces and simulating immersive visual effects like games, with insufficient research focusing on scalability.

In traffic sign detection and recognition (TSR), large-scale road scene datasets including traffic signs, such as Cityscapes [4] and Mapillary Vistas [5], as well as datasets specifically for traffic signs, such as GTSRB [6] and TT100K [7], have been developed, and improvements in recognition accuracy using these datasets have been reported. Furthermore, YOLO [8] and its derivatives [9] have enabled real-time object detection and class estimation. While previous research has shown significant improvements in recognition accuracy, most TSR studies focus on recognizing the location and class labels of markers rather than estimating highly accurate dimensions for 3D asset construction.

Traditional computer vision research has demonstrated that measurement information can be reconstructed from a single image if sufficient geometric features are available, as exemplified by single-view measurements [10] and vanishing point-based camera calibration [11]. Feature-based pipelines, along with multi-view approaches such as SfM [12,13] and SLAM [14], enable accurate measurement reconstruction when multiple images are available. However, these methods are not applicable when sufficient geometric features are lacking and only one image of the object is available.

Deep learning has significantly advanced monocular depth estimation from single images [15]. These approaches leverage large-scale datasets to produce visually plausible depth maps and 3D structures. However, they typically do not incorporate absolute scale information, and therefore accurate metric reconstruction remains challenging.

For example, 3D reconstruction from a single image has been explored using mesh-based approaches [16,17], voxel-based methods [18], and neural rendering techniques such as NeRF and its variants [19,20]. More recently, diffusion-based methods have enabled plausible 3D generation from a single image or even a text prompt [21,22]. However, these approaches are generally optimized for visual realism or the exploitation of category-level priors, and do not guarantee the metric accuracy required for digital twins that support measurement and simulation.

## 3. Method Overview

Fig.1 shows the processing flow of the proposed method. A single monocular image is input, and the object of the traffic sign, which is the target of detection, is detected in the Detection of target objects. Next, in Parts extraction, each component is extracted from the feature quantities of the detected target image using a parts detection model. In Size estimation, the size of each part is estimated, and there are two estimation methods depending on the conditions. The first method is to refer to a Parts data catalog which has past part size data and estimate the size based on the part with the highest confidence in size information from among the multiple detected parts. The second method is to refer to accumulated design information and extract design information that is close in size ratio to each part and apply that size. Finally, each part is 3D modeled based on the estimated dimensions, and the 3D object of the traffic sign is constructed by assembling the parts of these 3D models based on the design information. The characteristic processing shown in Fig.1, contents 3.1 to 3.3, will be explained in each section.

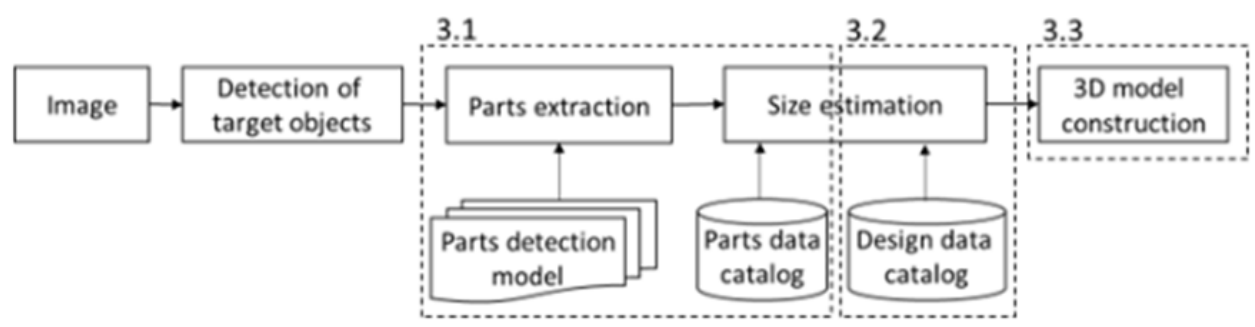

Fig.1 The workflow of the proposed method

### 3.1 Parts extraction for size estimation

Traffic signs are composite structures composed of multiple parts. This study proposes a method for estimating the size of other parts by individually detecting each part and using parts with known dimensions as a reference for scale estimation. The core of this method lies in estimating the scale based on known parts present in the image and propagating that information to other parts. As a preprocessing step for this, a process is performed to detect and extract each part from the image.

In this study, to construct a component detection model, we manually annotated each component region in traffic sign images and created training data by assigning class labels to each region. The components to be annotated were the sign board, the support post, the pole connecting the sign board and the support post, the arrow within the sign board, and the route number indicator region. For each component, the four vertices of the target region were manually obtained and converted into a bounding box format used for training the object detection model. For the arrow within the sign board and the route number indicator region, annotation was performed after perspective transformation to a planar image of the sign board viewed from the front. For example, the coordinates of the arrow axis and the route number indicator region shown in Fig. 2 were manually annotated using this procedure and used as training data for YOLO. The training data created consisted of over a dozen images of traffic signs, and a training data set was constructed targeting the components contained in each image. Using this dataset, the YOLO object detection model was trained to construct detection models for each component. As a result, a model with a detection accuracy of over 80% for the target components was constructed. It should be noted that this dataset is not intended for the construction of a large-scale general-purpose object detector but rather serves as an initial dataset for extracting standardized sign components as a reference standard for scale estimation and verifying the effectiveness of the proposed knowledge-driven size estimation framework.

Fig.2 shows an example of extracting arrows and symbols from an image of a traffic sign [23] using a part detection model, and associating known dimensions based on a parts data catalog with pixel dimensions on the image. To estimate part sizes, we propose a method of organizing known dimensional information that can be obtained in advance into a database called a "parts data catalog" and using it as a reference standard. The parts data catalog registers standardized dimensional information for each sign component, such as the arrow width included in traffic guidance signs, the vertical and horizontal dimensions of route number signs, and the character height. These dimensions are determined based on national and local government design guidelines for the purpose of ensuring safety and visibility and can therefore be used as reliable reference information. Specifically, for general road traffic signs, an arrow width of 150 mm, route number sign dimensions of 400 mm × 400 mm, and character height of 300 mm are registered. For highway traffic signs, an arrow width of 180 mm, route number sign dimensions of 500 mm × 500 mm, and character height of 500 mm are registered. In the proposed method, the pixel dimensions of the part area detected by the object detection model are compared with these actual dimensions registered in the part data catalog to estimate the scale corresponding to the part in the image. This allows standardized sign components to be used as a dimensional reference to estimate the size of the entire target sign, even when the image does not contain direct distance information.

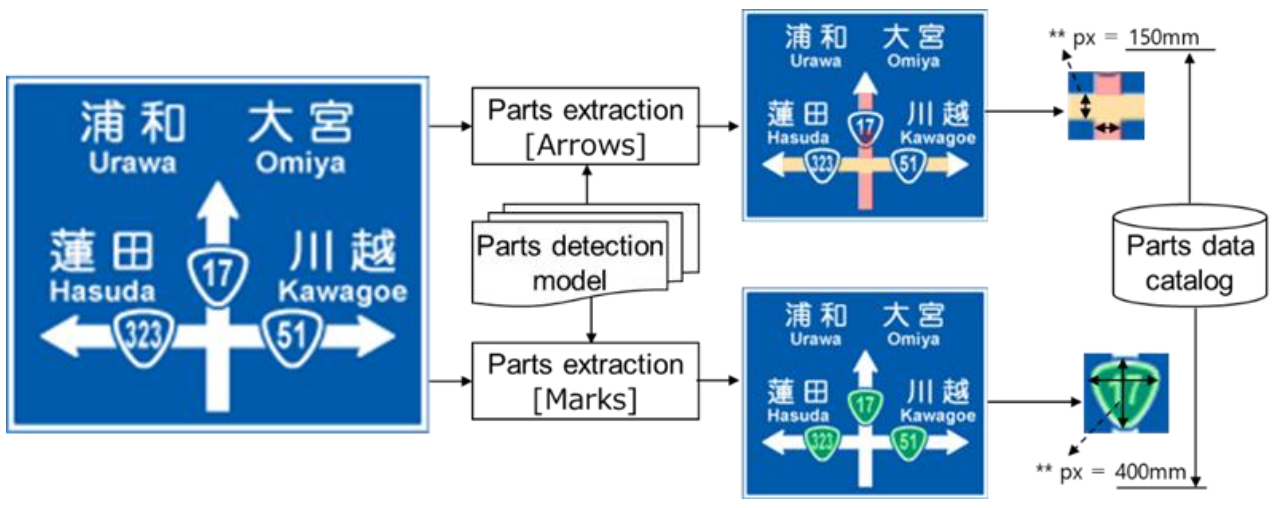

Fig.2 Parts extraction for size estimation

This method makes it possible to derive a standard for estimating scale within the sign surface. This standard is applicable to areas located on the same plane or at the same depth, but not to areas with different depths. This is because, in perspective projection, the size of an object on the image changes depending on its distance from the camera.

## 3.2 Knowledge-Driven Reference Selection

The part extraction based on the detection model used in Section 3.1 is susceptible to detection failures and misalignments of the detection area due to factors such as partial occlusion, image noise, and variations in lighting conditions. Such detection errors are unavoidable challenges in image recognition, and misalignments of the detection area significantly impact subsequent size estimation results. Therefore, in this study, when estimating the size of other parts based on the size estimated from an image, if the estimated value deviates significantly from the standard size, we consider the possibility that the reference detection result contains an error. In such cases, we propose a method for more reliable part size estimation by not using the reference size and instead referring to a large-scale design database of traffic signs.

The design data catalog shown in Fig.3 organizes and stores information about the sign structure extracted from the design drawings[24] as prior knowledge. This catalog is structured in a format where one record is associated with each sign type or design specification, and each record registers structural parameters such as sign width, sign height, support column diameter, support column length, pole diameter, pole length, distance from the end of the sign to the center of the support column, and distance between poles. Furthermore, analysis of this design information confirmed that a certain proportional relationship exists between multiple structural parameters, such as the ratio of column diameter to column length, the ratio of plate width or height to column length, and the relationship between pole length and sign width, even between signs of different sizes. In this study, we refer to a design data catalog that includes these dimensional ratios and arrangement relationships, and evaluate whether the dimensional ratios obtained from the detection results of multiple parts are consistent with the design values in the catalog. Specifically, for each detected combination of column diameter, sign dimensions, and pole position, the ratio of column diameter to signboard dimensions, the ratio of column length to signboard height, and the ratio of pole-to-pole distance to signboard width are calculated. If these ratios match the design specifications in the catalog, the detection result is determined to be based on the corresponding design specifications. Then, by referring to the structural parameters registered in the corresponding design record, the size of each part, including undetected or uncertain components, is estimated. Since the design data is designed to meet safety standards such as preventing tipping, the estimation results based on it have the advantage of not deviating significantly from actual structures.

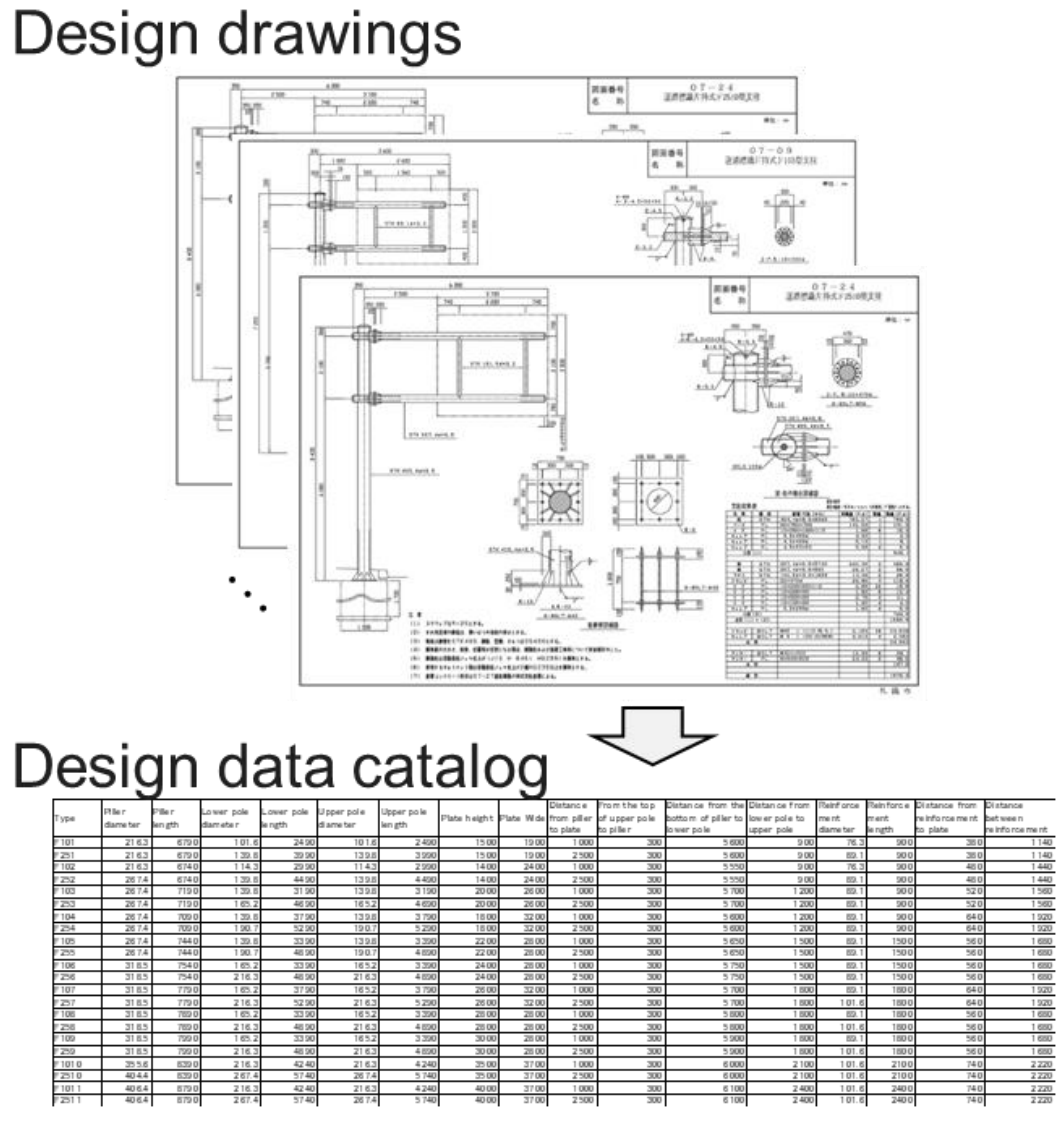

Fig.3 Design data catalog

This method functions as a fail-safe mechanism for size estimation in Section 3.1. Input image quality varies, and detection accuracy may decrease. However, by introducing this mechanism, it becomes possible to reconstruct an object with dimensions close to those of a real structure, even when incomplete part detection occurs.

### 3.3 3D Object Generation by Reconstructing Parts

In recent years, AI-based methods for generating 3D objects directly from a single 2D image have attracted attention. These approaches utilize generative models such as diffusion models and Neural Radiance Fields (NeRF) and are trained using a large amount of 3D assets and images rendered from them. As a result, it becomes possible to generate consistent and reasonable 3D shapes from input images based on prior distributions of 3D shapes acquired during the training process.

However, the generated 3D models are generally represented as a single mesh, meaning that structural separation of individual components is not considered. As a result, it becomes difficult to individually replace or edit specific components such as signs, or to modify each component, which limits their applicability and reusability.

In the proposed method, as shown in Fig.4, after determining the size of each component, each is generated as an independent 3D object, and the overall 3D structure is reconstructed by integrating these objects.

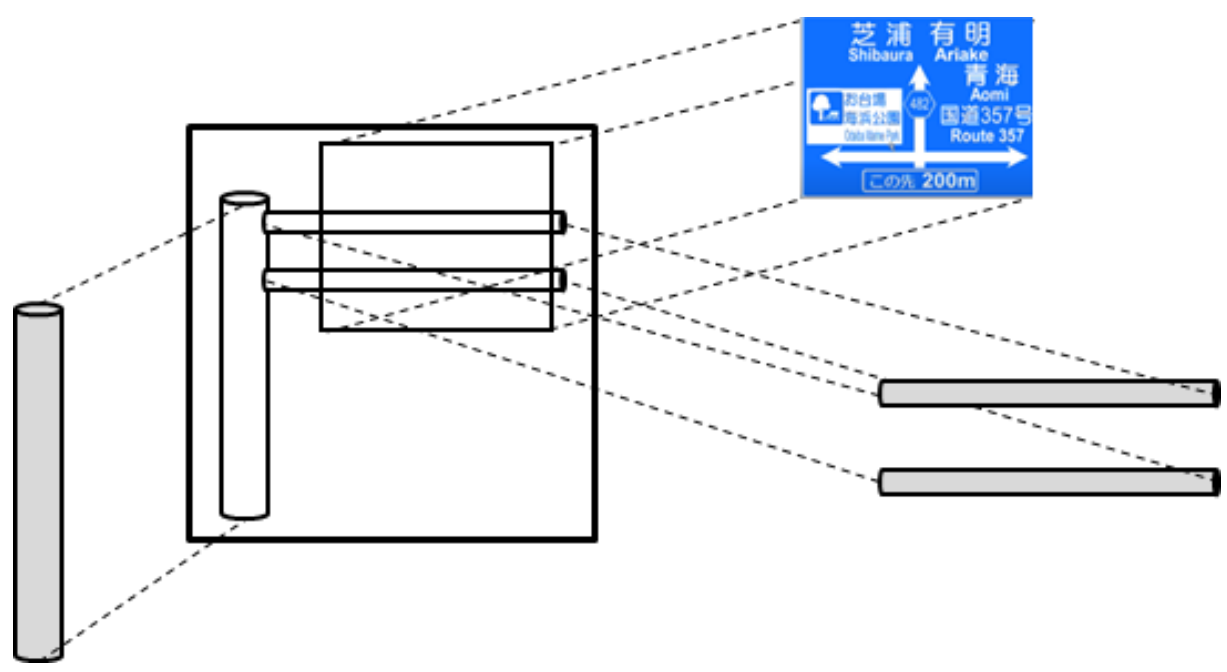

Fig.4 Assemble parts to create 3D objects

Basic geometric primitives such as cylinders and planes are parametrically generated and placed based on the design information. This method facilitates editing operations such as changing the shape and adjusting parameters at the component level and allows for flexible adaptation to a variety of simulation scenarios.

## 4. Contributions

The contributions of this work are summarized as follows:

1. A knowledge-driven framework for estimating the actual scale of traffic signs from monocular images was proposed, enabling the generation of 3D assets close to real dimensions even in situations where point cloud acquisition of high-altitude equipment using LiDAR or stereo cameras is difficult.
2. A size estimation method was proposed that decomposes traffic signs into their constituent parts for detection and reference's part data catalogs and design data catalogs, achieving fail-safe dimensional estimation based on parts with known dimensions and dimensional ratios between parts.
3. A method was proposed to generate 3D objects at the part level based on the estimated dimensions of each part and reconstruct them according to design information, improving editability, reusability in the digital twin space, and scale consistency for in-vehicle camera verification.

## 5. Discussion

This research is characterized by its ability to impart not only visual similarity but also dimensional consistency with the real world to 3D assets generated from monocular images. In automotive camera verification using digital twins, dimensional consistency is a crucial requirement because the scale of the object directly affects recognition performance.

The proposed method links detection results at the component level with design data, clearly demonstrating the basis for dimensional estimation and enabling error analysis. This is a characteristic not adequately addressed by existing 3D generation methods based on visual appearance. On the other hand, the performance of this method depends on the accuracy of part detection and the comprehensiveness of catalog information. Future challenges include quantitative evaluation based on diverse real-world data and the formulation of estimation uncertainty.

This method primarily performs scale estimation for components located within the sign surface or at the same depth. For components located at different depths, the same pixel scale cannot be directly applied because of perspective projection. To address this problem, it is necessary to estimate the depth of each component using a stereo camera or similar method, and then calculate the actual dimensions based on the distance from the camera using a pinhole camera model. In other words, it is effective to perform scale correction that takes depth differences into account, using the size of the part in the image, the focal length, and the distance to the target part. However, this paper mainly focuses on the effectiveness of size estimation based on monocular images and design knowledge, and verification integrating depth estimation for each part will be a task for the future.

## 6. Limitations and Future Work

The current framework assumes the availability of objects specific to structural knowledge and is therefore best suited to domains with standardized designs.

Extending the approach to less constrained object categories and formalizing uncertainty measures for reference selection are promising directions for future work.

## 7. Conclusion

This paper focuses on traffic guidance signs installed throughout Japan, whose content and size vary depending on installation conditions. These signs are often installed at high altitudes, making direct on-site measurement difficult; therefore, scale estimation from monocular images is required.

This research proposes a knowledge-driven framework that enables consistent dimensional estimation even when direct scale information is unavailable or incomplete. Specifically, it estimates the overall size by decomposing the object into its constituent elements and propagating the scale by referencing publicly available partial dimensional information for each part.

Furthermore, this method utilizes design information as a fail-safe mechanism in addition to image detection results, resulting in high robustness against detection errors and partial observations. In addition, the reconstruction method at the part level enables the generation of editable 3D objects, making it suitable for digital twin environments requiring diverse verification.

Future challenges include developing design knowledge to accommodate the expansion of target objects and evaluation in large-scale simulation environments for autonomous driving verification.

**Acknowledgments**

The authors wish to thank all the subjects who participated in this study.

The author, Mr. Akai*, was researching the content of this paper in the same department but is now affiliated with the following organization.

*Astemo, Ltd. Technology Development Functional Division Vehicle Integrated Control Technology Development Dept. 7-1, Onna 4-chome, Atsugi-shi, Kanagawa, 243-8510 Japan